\documentclass{article}

\usepackage{microtype}
\usepackage{graphicx}
\usepackage{subfigure}
\usepackage{booktabs} % for professional tables
\usepackage{tikz}
\usetikzlibrary{matrix}
\usepackage{hyperref}

\usepackage[archival]{icml2023}

\usepackage{amsmath}
\usepackage{amssymb}
\usepackage{mathtools}
\usepackage{amsthm}

\usepackage[capitalize,noabbrev]{cleveref}

\theoremstyle{plain}
\newtheorem{theorem}{Theorem}[section]

\theoremstyle{definition}
\newtheorem{definition}[theorem]{Definition}

\theoremstyle{remark}

\usepackage[textsize=tiny]{todonotes}

\icmltitlerunning{Learning Lie Group Symmetry Transformations with Neural Networks}

\begin{document}

\twocolumn[
\icmltitle{Learning Lie Group Symmetry Transformations with Neural Networks}



\icmlsetsymbol{equal}{*}

\begin{icmlauthorlist}
\icmlauthor{Alex Gabel}{equal,yyy}
\icmlauthor{Victoria Klein}{equal,sch}
\icmlauthor{Riccardo Valperga}{equal,yyy}
\icmlauthor{Jeroen S. W. Lamb}{sch}
\icmlauthor{Kevin Webster}{sch}
\icmlauthor{Rick Quax}{xxx}
\icmlauthor{Efstratios Gavves}{yyy}
\end{icmlauthorlist}

\icmlaffiliation{sch}{Department of Mathematics, Imperial College London}
\icmlaffiliation{yyy}{VIS Lab (Institute of Informatics), University of Amsterdam}
\icmlaffiliation{xxx}{CSL (Institute of Informatics), University of Amsterdam}

\icmlcorrespondingauthor{Victoria Klein}{victoria.klein18@imperial.ac.uk}
\icmlcorrespondingauthor{Alex Gabel}{a.gabel@uva.nl}
\icmlcorrespondingauthor{Riccardo Valperga}{r.valperga@uva.nl}

\icmlkeywords{Machine Learning, ICML}

\vskip 0.3in
]



\printAffiliationsAndNotice{\icmlEqualContribution} % otherwise use the standard text.

\begin{abstract}
The problem of detecting and quantifying the presence of symmetries in datasets is useful for model selection, generative modeling, and data analysis, amongst others. While existing methods for hard-coding transformations in neural networks require prior knowledge of the symmetries of the task at hand, this work focuses on discovering and characterizing unknown symmetries present in the dataset, namely, Lie group symmetry transformations beyond the traditional ones usually considered in the field (rotation, scaling, and translation). Specifically, we consider a scenario in which a dataset has been transformed by a one-parameter subgroup of transformations with different parameter values for each data point. Our goal is to characterize the transformation group and the distribution of the parameter values. The results showcase the effectiveness of the approach in both these settings.
\end{abstract}

% \begin{itemize}
% \item \textbf{New to this year}: If your paper has appendices, submit the appendix together with the main body and the references \textbf{as a single file}.
% \item Page limit: The main body of the paper has to be fitted to 8 pages, excluding references and appendices; the space for the latter two is not limited. For the final version of the paper, authors can add one extra page to the main body.
% \item Place figure captions \emph{under} the figure (and omit titles from inside
%     the graphic file itself). Place table captions \emph{over} the table.
% \item References must include page numbers whenever possible and be as complete
%     as possible. Place multiple citations in chronological order.
% \item The paper abstract should begin in the left column, 0.4~inches below the final
% address. The heading `Abstract' should be centered, bold, and in 11~point type.
% The abstract body should use 10~point type, with a vertical spacing of
% 11~points, and should be indented 0.25~inches more than normal on left-hand and
% right-hand margins. Insert 0.4~inches of blank space after the body. One paragraph and roughly
%     4--6 sentences. The title should have content words capitalized.
% \item Please use no more than three levels of headings.
% \item Do not indent the first line of a given
% paragraph, but insert a blank line between succeeding ones.
% \end{itemize}

\section{Introduction}\label{introduction}
It has been shown that restricting the hypothesis space of functions that neural networks are able to approximate  using known properties of data improves performance in a variety of tasks \citep{worrall2019deep, cohen2018spherical, weiler20183d, zaheer2017deep, cohen2016group}. The field of Deep Learning has produced a prolific amount of work in this direction, providing practical parameterizations of function spaces with the desired properties that are also universal approximators of the target functions \citep{yarotsky2022universal}. In physics and, more specifically, time-series forecasting of dynamical systems, symmetries are ubiquitous and laws of motion are often symmetric with respect to various transformations such as rotations and translations, while transformations that preserve solutions of equations of motions are in one way or another associated with conserved quantities \citep{Noether1918}. In computer vision, successful neural network architectures are often invariant with respect to transformations that preserve the perceived object identity as well as all pattern information, such as translation, rotation and scaling. Many of these transformations are smooth and differentiable, and thus belong to the family of Lie groups, which is the class of symmetries we deal with in this work.
\begin{figure}[t]
    \centering
 \includegraphics[width=\linewidth]{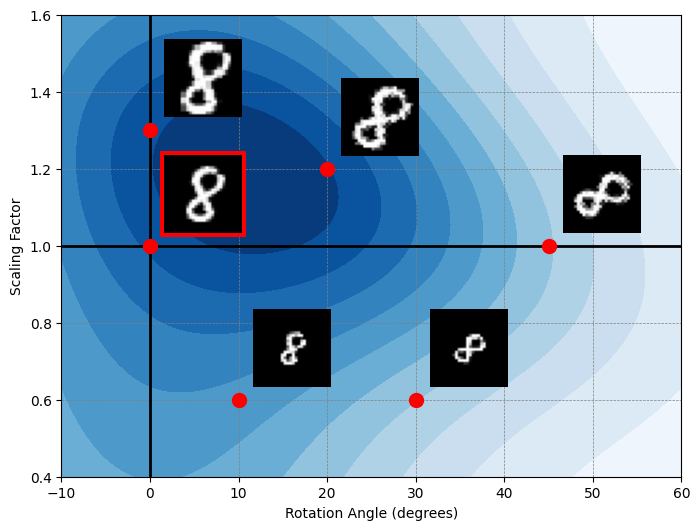}
    \caption{The distribution of transformations in a toy dataset that correspond to the Lie groups of rotation and  (isotropic) scaling, given in terms of the parameters degree and scaling factor respectively; crucially, these groups are differentiable and can be (locally)  decomposed into one-parameter subgroups.}
    \label{fig:distribution}
\end{figure}

Although methods that hard-code transformations are capable of state-of-the-art performance in various tasks, they all require prior knowledge about symmetries in order to restrict the function space of a neural network. A broad class of, a priori unknown, transformations come into play in the context of modelling dynamical systems and in applications to physics. On the other hand, in vision tasks, identity-preserving transformations are often known beforehand. Despite this, these transformations are expressed differently by different datasets. As a result, algorithms for not only \emph{discovering} unknown symmetries but also \emph{quantifying} the presence of specific transformations in a given dataset, may play a crucial role in informing model selection for scientific discovery or computer vision, by identifying and describing physical systems through their symmetries and selecting models that are invariant or equivariant with respect to only those symmetries that are \emph{actually} present in the dataset under consideration.

In this work, we address the problem of qualitatively and quantitatively detecting the presence of symmetries with respect to one-parameter subgroups within a given dataset (see Figure \ref{fig:distribution}). In particular, let $\phi(t)$ be a one parameter subgroup of transformations. We consider the scenario in which a dataset $\{x_i\}_{i=1}^{N}$ has been acted on by $\phi(t)$, with a \emph{different} value of the parameter $t$ for every point $x_i$. Our goal is to characterise the group of transformations $\phi(t)$, as well as the \emph{distribution} from which the parameters $t$ have been sampled. We propose two models: a naive approach that successfully manages to identify the underlying one-parameter subgroup, and an autoencoder model that learns transformations of a one-parameter subgroup in the latent space and is capable of extracting the overall shape of the $t$-distributions. The cost of the latter is that the one-parameter subgroup in the latent space is not necessarily identical to that in pixel space. The work is structured as follows: Section \ref{sec:theory} introduces some basic tools from Lie group theory; Section \ref{sec:method} outlines the method; Section \ref{sec:related} provides an overview of the existing methods that are related to our own; and lastly, results are shown in Section \ref{sec:experiments}.

\section{Background}
\label{sec:theory}

The theoretical underpinnings of symmetries or invariance can be described using group theory \citep{fulton1991representation}. In particular, we present the necessary theory of one-parameter subgroups \citep{olver1993applications} on which our method is based, following the logic of \citet{oliveri}.

\subsection{One-parameter subgroups}
\label{subsec:one-parameter}
We focus on learning invariances with respect to one-parameter subgroups of a Lie group $G$, which offer a natural way to describe continuous symmetries or invariances of functions on vector spaces.

\begin{definition}
    \label{def:liegroupoft}
    A \textbf{one-parameter subgroup} of $G$ is a differentiable homomorphism $\phi:\mathbb{R}\to G$, more precisely, such that $\phi(t+s) = \phi(t)\phi(s)$ for all $t,s\in\mathbb{R}$. 
\end{definition}

Let the action of $\phi$ on the vector space $X\subset\mathbb{R}^n$ be a transformation $T:X\times\mathbb{R}\to X$ that is continuous in $x\in X$ and $t\in\mathbb{R}$. Because of continuity, for sufficiently small $t$ and some fixed $x\in X$, the action is given by
\begin{equation}
\label{eq:bla}
    T(x,t)\approx x+tA(x)\;\text{where}\; A(x):=\frac{\partial T(x,t)}{\partial t}\Bigg|_{t=0}.
\end{equation}
Note that this is equivalent to taking a first-order Taylor expansion in $t$ around  $t=0$.

\subsection{Generators}

In general, we can use $A(x)$ in \eqref{eq:bla} to construct what is known as the \textbf{generator} of a one-parameter subgroup $\phi$ of a Lie group $G$, that in turn will characterise an ordinary differential equation, the solution to which coincides with the action $T$ on $X$. 

Let $C^\infty(X)$ be the space of smooth functions from $X$ to $X$. The generator of $\phi$ is defined as a linear differential operator $L:C^\infty(X)\to C^\infty(X)$ such that
\begin{equation}
\label{eq:generator}
    L = \sum_{i=0}^n(A(x))_i\frac{\partial}{\partial x_i}
\end{equation}
describing the vector field of the infinitesimal increment $A(x)t$ in \eqref{eq:bla}, where $\partial/\partial x_i$ are the unit vectors of $X$ in the coordinate directions for $i=1,\hdots,n$.  It can be shown \citep{olver1993applications} that, for a fixed $x\in X$, that $T(x,t)$ is the solution to the ordinary differential equation
\begin{equation}
    \label{eq:flow}
    \frac{dT(x,t)}{dt}=LT(x,t)\quad\text{where}\quad T(x,0)=x.
\end{equation}

The solution to \eqref{eq:flow} is the exponential $T(x,t)=e^{tL}x$ where
\begin{equation}
    \label{eq:exponential}
    e^{tL}:= \sum_{k=0}^\infty \frac{(tL)^k}{k!},
\end{equation}
where $L^k$ is the operator $L$ applied $k$ times iteratively.

For a one-parameter subgroup $\phi$ of a matrix Lie group $G\subset GL(n,\mathbb{R})$ and a fixed $x\in X$, it can be shown \citep{olver1993applications} that there exists a unique matrix $A\in\mathbb{R}^{n\times n}$ such that $A(x)=Ax$. This is a more restrictive approach as groups such as translations cannot be written as a matrix multiplication.

\begin{figure*}[h]
    \centering
    \includegraphics[width=0.9\linewidth]{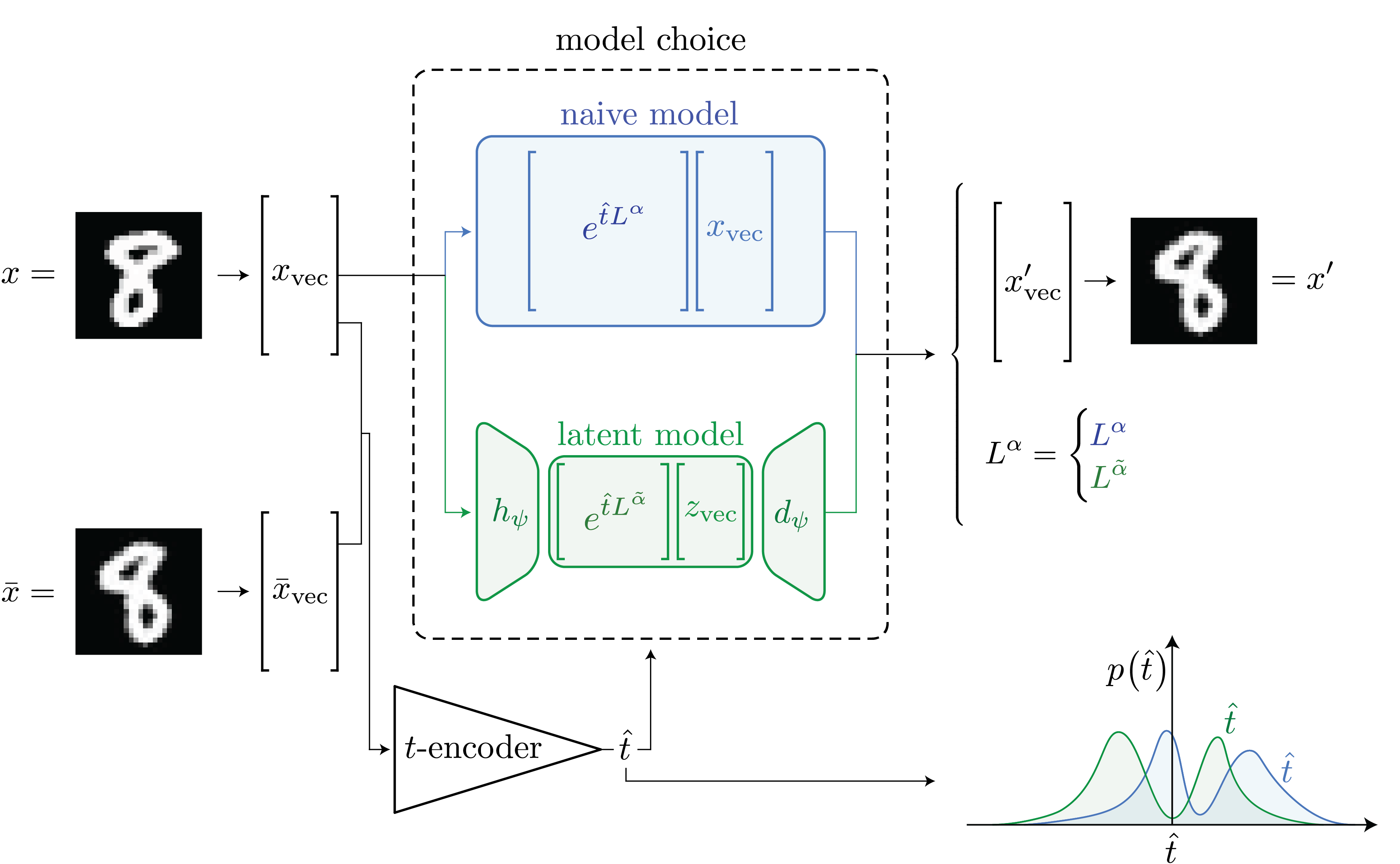}
    \caption{Model architecture.}
    \label{fig:architecture}
\end{figure*}

\section{Method}\label{sec:method}
As in \citet{rao1998learning, sanborn2022bispectral, dehmamy2021automatic} the semi-supervised symmetry detection setting that we consider consists of learning the generator $L$ of a one-parameter subgroup $\phi$ from pairs of observations of the form $\left\{\left(x_i, \bar{x} = T(x_i, t_i)\right)\right\}_{i=1}^{N}$, where $N$ is the number of observations and each $t_i\in\mathbb{R}$ is drawn from some unknown distribution $p(t)$. Not only do we attempt to learn the generator $L$, but also the unknown distribution $p(t)$ of the parameters $\{t_i\}_{i=1}^N$.

\subsection{Parametrisation of the generator}
\label{subsec:paramgen}
Deciding how to parametrise $L$ has an effect on the structure of the model and ultimately on what one-parameter subgroups we are able to learn. For simplicity, consider one-parameter subgroups acting on $X\subset\mathbb{R}^2$, although this operator can be defined for higher-dimensional vector spaces. The generator $L$ of $\phi$ is given as in Eq. \eqref{eq:generator} and we parametrise $A(x,y)$ as a linear operator in the basis $\{1,x,y\}$ with a coefficient matrix $A=\alpha\in\mathbb{R}^{2\times 3}$, giving
\begin{align}
    \label{eq:modelgenerator}
    \begin{split}
        L^{\alpha}&:=
        (\alpha_{11} + \alpha_{12}x + \alpha_{23}y)\frac{\partial}{\partial x} \\
        &+(\alpha_{12} + \alpha_{22}x + \alpha_{23}y)\frac{\partial}{\partial y}.
    \end{split}
\end{align}
 In this particular basis, for different values of $\alpha$, the generator $L^\alpha$ is able to express one-parameter sub-groups of the affine group. This includes the ``traditional" symmetries that are usually considered (translation,  rotation, and isotropic scaling) and all other affine transformations\footnote{Alternatively, the constant terms can be thought of as the drift terms (i.e. translation) and the four others can be arranged into a diffusion matrix.}. This can be generalized to any functional form of the generator by augmenting the basis accordingly.

 % It remains to interpret the generator $L^\alpha$ in \eqref{eq:modelgenerator}. However, the transformation $\hat{T}$ in \eqref{eq:modelapprox} is not well-defined on the space $M\subset\mathbb{R}^{28\times 28}$ representing $\mathbb{R}^2$ because it requires a notion of infinitesimal increments ${\partial}/{\partial x}$ and ${\partial}/{\partial y}$ on $M$ and a representation of $I_i$ in some vector space.

\subsection{Discretisation and interpolation}
\label{subsec:discretegen}
The generator $L^\alpha$ is constructed as an operator that acts on a function $f:\mathbb{R}^2\to\mathbb{R}$, given, in practice, by $I\in\mathbb{R}^{n\times n}$ such that $I_{ij}=f(i, j)$ are evaluations of $f$ on a regularly-sampled $n\times n$ grid $M$ of points $M_{ij}=(i, j)\in\mathbb{R}^2$. We then vectorise $I$, obtaining a point in a vector space $\tilde{I}\in\mathbb{R}^{n^2}$ such that $\tilde{I}_{i+j}:=I_{ij}$ and construct the matrix operator $L^\alpha\in\mathbb{R}^{n^2\times n^2}$ as
\begin{align}
    \label{eq:modelgenerator2}
    \begin{split}
        L^{\alpha}&:=
        (\alpha_{11} + \alpha_{12}X_x + \alpha_{13}X_y)\frac{\partial}{\partial X_x}\\
        &+(\alpha_{21} + \alpha_{22}X_x + \alpha_{23}X_y)\frac{\partial}{\partial X_y},
    \end{split}
\end{align}
acting on $\tilde{I}$, where $X_x\in\mathbb{R}^{n^2\times n^2}$ and $X_y\in\mathbb{R}^{n^2\times n^2}$ are such that $(X_{x})_{ij}:=i$ and $(X_{y})_{ij}:=j$, while ${\partial}/{\partial X_x}$ and ${\partial}/{\partial X_y}$ are also matrix operators in $\mathbb{R}^{n^2\times n^2}$.
The exponential in \eqref{eq:exponential} and the action $T$ coincides with the matrix exponential.

In order to define ${\partial}/{\partial X_x}$ and ${\partial}/{\partial X_y}$ as operators that transform by infinitesimal amounts at discrete locations, we require an interpolation function. The Shannon-Whittaker theorem \citep{marks2012introduction} states that any square-integrable, piecewise continuous function that is band-limited in the frequency domain can be reconstructed from its discrete samples if they are sufficiently close and equally spaced. For sake of interpolations, we will also assume that the function is periodic.

\paragraph{Interpolation: 1D}
In the case where $M$ is a discrete set of $n$ points in 1D, we have that $I(i + n) = I(i)$ for all $i=1,\hdots,n$ samples. Shannon-Whittaker interpolation reconstructs the signal for all $x\in\mathbb{R}$ as 
\begin{equation}
    \label{eq:shannon}
    \begin{split}
            &I(x) = \sum_{i=0}^{n-1}I(i)Q(x-i), \quad \text{where}\\
            &Q(x) =  \frac{1}{n}\left[ 1+2\sum_{p=1}^{n/2-1}\cos\left(\frac{2\pi p x}{n}\right)\right]
    \end{split}
\end{equation}
Differentiating $Q$ with respect to $x$ and evaluating it at every $x_i\in M$ gives an analytic expression for a vector field in $\mathbb{R}^n$, describing continuous changes in $x$ at all $n$ points \citep{rao1998learning}. This is precisely what ${\partial}/{\partial x}$ or ${\partial}/{\partial y}$ in \eqref{eq:modelgenerator} are.

\paragraph{Interpolation: 2D}
In the case where $M$ is a grid of $n\times n$ points in 2D, we construct the $n\times n$ matrices of the partial derivatives of $Q$ with respect to $x$ and $y$, analogously to the 1D case, stacking them to construct the ${n^2\times n^2}$ block diagonal matrices  ${\partial}/{\partial X_x}$ and ${\partial}/{\partial X_y}$. It is worth noting that alternative interpolation techniques can be used to obtain the operators and the method does not depend on any specific one.

Two different architectures, the main model and the latent model, are proposed to learn $L^\alpha$ and, in doing so, the action $T$.

\subsubsection{Naive model}
\label{subsubsec:mainmodel}
The coefficients $\alpha$ of $L^\alpha$ are approximated by fixed coefficients that are shared across the dataset, while the parameter $t_i$ is approximated by $\hat{t}_i$ that depends on the input pair $(x_i, \bar{x}_i)$. We learn
\begin{enumerate}
    \item the coefficients $\alpha\in\mathbb{R}^{2\times 3}$ of the generator $L^{\alpha}$ and
    \item the parameters $\theta$ of an MLP $f_{\theta}$ that returns $f_{\theta}(x_i, \bar{x}_i)=:\hat{t}_i$ as a function of every input pair,
\end{enumerate}
such that the solution to \eqref{eq:flow} for $L^{\alpha}$ is approximated by
\begin{equation}
    \label{eq:modelapprox}
    \hat{T}(x_i, \bar{x}_i):=e^{f_{\theta}(x_i, \bar{x}_i)L^{\alpha}}\,x_i.
\end{equation}
The model objective is then given by the reconstruction loss
\begin{equation}
    \label{eq:mainloss}
    \mathcal{L}_T(x_i, \bar{x}_i)=||\hat{T}_\phi(x_i, \bar{x}_i)-\bar{x}_i||^2.
\end{equation}

% The exponential in \eqref{eq:modelapprox} is in need of motivation and not well-defined. The use of the exponential is motivated by the form of the ordinary differential equation of the flow in \eqref{eq:flow}. In the case where $(x_i, \bar{x}_i)\in\mathbb{R}^n \times \mathbb{R}^{n}$, the unit vectors $\partial/\partial x$ in $L^{\alpha}$ are well-defined continuous operators and these unit vectors are omitted when evaluating \eqref{eq:modelapprox} in the forward pass of the model, taking $L^{\alpha}=\alpha x$. Then the exponential in \eqref{eq:modelapprox} is well-defined and coincides with the matrix exponential.

% If however, as is the case in experiments in Section \ref{sec:experiments}, the inputs $(x_i, \bar{x}_i)$ lie in a space on which the action of $G$ is not well defined (e.g. images represented by pixel matrices) then the inputs are mapped to a vector space in $\mathbb{R}^n$ (e.g. images are vectorised). Correspondingly, unit vectors $\partial/\partial x\in\mathbb{R}^{n\times n}$ are defined (e.g. through interpolation) as matrices that describe infinitesimal changes with respect to each coordinate in the new vector-valued inputs. Section \ref{subsec:interpolation} outlines this procedure for when the inputs are image signals.

%%%%%%%%%%%%%%%%%%%%%%%%%%%%%
\subsubsection{Latent model}
\label{subsubsec:latentmodel}
%%%%%%%%%%%%%%%%%%%%%%%%%%%%%

While the model described above will prove to work sufficiently well for learning the coefficients $\alpha$ of $L^{\alpha}$, the matrix exponential function in $\hat{T}$ in \eqref{eq:modelapprox} can be costly to compute and difficult to optimise in high dimensions; consider that the cost of the matrix exponential in a single forward pass is roughly $O(n^3)$ using the algorithm of \citet{al2010new}.
% when the inputs to the model are images in $\mathbb{R}^{28\times 28}$ that are vectorised to $\mathbb{R}^{784}$,

As a result, a different version of the model is proposed that incorporates an autoencoder for reducing dimension. The concept remains the same, but $x_i$ is now mapped to some latent space $Z\subset\mathbb{R}^{n_Z}$ for $n_Z\ll n$, such that the exponential is taken in a significantly lower dimension. This is done by an encoder $h_\psi:X\to Z$ and a decoder $d_\psi:Z\to X$ such that $z_i=h_\psi(x_i)$ and $x_i\approx d_\psi(z_i)$.

We learn
\begin{enumerate}
    \item the parameters $\psi$ of an MLP autoencoder,
    \item the coefficients $\tilde{\alpha}\in\mathbb{R}^{2\times 3}$ of the generator $L^{\tilde{\alpha}}$ for a one-parameter subgroup $\phi_Z$ acting on the latent space $Z$,
    \item the parameters $\theta$ of an MLP $f_\theta$ that returns $f_{\theta}(x_i, \bar{x}_i)=:\hat{t}_{i}$ as a function of every original input pair $(x_i, \bar{x}_i)$,
\end{enumerate}
such that the solution to \eqref{eq:flow} for $L^{\alpha}$, the generator in the original space, is approximated by
\begin{equation}
    \label{eq:latentmodelapprox}
    \hat{T}^Z(x_i, \bar{x}_i)=d_\psi(e^{f_{\theta}(x_i, \bar{x}_i)L^{\tilde{\alpha}}}h_\psi(x_i)).
\end{equation}

\begin{figure}[h]
\centering
\begin{tikzpicture}[>=stealth, font=\Large, line width=1.5pt]
  \node[font=\huge] (A) at (0,0) {$X$};
  \node[font=\huge] (B) at (5,0) {$X$};
  \node[font=\huge] (C) at (0,-3) {$Z$};
  \node[font=\huge] (D) at (5,-3) {$Z$};
  \draw[->] (A) -- node[above] {$T(\cdot, t)$} (B);
  \draw[->] (C) -- node[below] {$e^{f_\theta(\cdot, \cdot)L^{\tilde{\alpha}}}$} (D);
  \draw[->] (A) -- node[left] {$h_\psi$} (C);
  \draw[<-] (B) -- node[right] {$d_\psi$} (D);
  %\node at (2,-1) {\large\textbf{Equivariance}};
\end{tikzpicture}
\caption{The commuting diagram enforced by the objective function in the latent model: $T(x,t) \approx d_\psi(e^{f_{\theta}(x_i, \bar{x}_i)L^{\tilde{\alpha}}} h_\psi(x))$.}
\label{fig:diagram}
\end{figure}
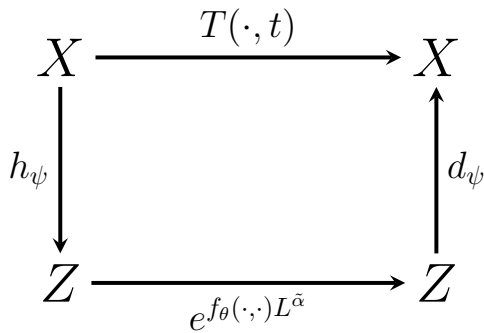

It is important to note that enforcing good reconstruction of the autoencoder alone does not enforce the commutativity of the diagram in Figure \ref{fig:diagram}. To make it commutative, we use an objective that is a weighted sum of multiple terms. A simple reconstruction term for the autoencoder on each input example
\begin{equation}
    \label{eq:reconloss}
    \mathcal{L}_R(x_i):=||d_\psi(h_\psi(x_i))-x_i||^2,
\end{equation}
a transformation-reconstruction term in the original space
\begin{equation}
    \label{eq:}
    \mathcal{L}^X_T(x_i, \bar{x}_i):=||\hat{T}^Z_\phi(x_i, \bar{x}_i)-\bar{x}_i||^2,
\end{equation}
a transformation-reconstruction term in the latent space
\begin{equation}
    \label{eq:}
    \mathcal{L}^Z_T(x_i, \bar{x}_i):=||e^{f_{\theta}(x_i, \bar{x}_i)L^{\tilde{\alpha}}}h_\psi(x_i) - h_\psi(\bar{x}_i)||^2,
\end{equation}
and a Lasso term on the generator coefficients $\tilde{\alpha}$. The overall loss of the latent model is
\begin{align}
\label{eq:latentloss}
\begin{split}
    \mathcal{L}(x_i, \bar{x}_i) &= \lambda_{R}(\mathcal{L}_R(x_i)+\mathcal{L}_R(\bar{x}_i))\\
    &+\,\lambda_X\mathcal{L}^X_T(x_i, \bar{x}_i)+\lambda_Z\mathcal{L}^Z_T(x_i, \bar{x}_i)\\
    &+\lambda_{L}||\mathbf{\tilde{\alpha}}||^{2},
\end{split}
\end{align}
where $\lambda_R,\lambda_X,\lambda_Z, \lambda_L\in\mathbb{R}$ are treated as 
hyperparameters.

\paragraph{Recovering the group}
It is important to note that the one-parameter subgroup corresponding to the generator $L^{\tilde{\alpha}}$ and the generator $L^{\alpha}$ are \emph{not} necessarily the same; $L^{\tilde{\alpha}}$ is the generator corresponding to some action on $X$ of a one-parameter subgroup $\phi$, while $L^{\alpha}$ is a different generator corresponding to some action on $Z$ of a different one-parameter subgroup $\phi_Z$. 

% Analogously, the values of $\hat{t}_Z$ approximate the parameters $t_Z$ corresponding to $L^{\tilde{\alpha}}$ \emph{not} the parameters $t$ corresponding to $L^{\alpha}$.
% However, the \emph{whole} transformation $\hat{T}^Z$ in \eqref{eq:latentmodelapprox} approximates the action on of some one-parameter subgroup on $X$, whose generator can be recovered by the derivative with respect to $t$ at $t=0$, as shown in \eqref{eq:taylor2}. Therefore, we define the approximation of this generator as 
% \begin{equation}
%     \label{eq:revgenerator}
%     A_Z(z):=\frac{\partial \hat{T}^Z(z,\hat{t}_Z)}{\partial \hat{t}_Z}\Bigg|_{\hat{t}_Z=0},
% \end{equation}
% which itself can be approximated numerically after training.

\subsection{Uniqueness}
For both the naive model in Section \ref{subsubsec:mainmodel} and the latent model in \ref{subsubsec:latentmodel}, the approximations $\hat{t}_i$ for the values of the parameters $t_i$ require interpretation.
Both models parameterise $\hat{T}$ or $\hat{T}^Z$ with the products $\hat{t}_iL^{\alpha}$ or $\hat{t}_iL^{\tilde{\alpha}}$ respectively, where $\hat{t}_i = f_{\theta}(x_i, \bar{x}_i)$. While both the values of $\hat{t}_iL^{\alpha}$ and $\hat{t}_iL^{\tilde{\alpha}}$ are unique for a given action on $X$ and $Z$ respectively, their decomposition is only unique up to a constant. Therefore, $L^{\alpha}$ or $L^{\tilde{\alpha}}$ and $\hat{t}$ approximate the generators and the parameter respectively up to a constant. Consequently, the one-parameter subgroup $\phi$ can only be deduced by the values of the individual coefficients in $\alpha$ \emph{relative to one another}, as opposed to in absolute, likewise for $\phi_Z$ and $\tilde{\alpha}$ . We therefore recover a scaled approximation for the distribution of $\hat{t}_i$.

\subsection{The most general setting}\label{sec:generalsetting}
Suppose we are given a labelled dataset $\mathcal{D} = \left\{(x_i, c_i)\right\}_{i=1}^{N}$ and a one-parameter subgroup $\phi$. Then we call $\mathcal{D}$ \textit{symmetric} or \textit{invariant with respect to} $\phi$ if the action of $\phi$ preserves the object identity of the data points, where by object identity we mean any property of the data that we might be interested in. For example, in the case of MNIST handwritten digits, rigid transformations preserve their labels \footnote{With the exception of the number '9' that, if rotated 180 degrees becomes a '6'.} and therefore, can be considered symmetries of the dataset. Now suppose that every $x_i$ in $\mathcal{D}$ is acted on with a one-parameter subgroup $\phi_{t}$ to get $T\mathcal{D} = \left\{(T(x_i, t_i), c_i)\right\}_{i=1}^{N}$. The most general, fully unsupervised symmetry detection setting consists of learning $\phi$, and characterize the distribution of the parameter $t$ from just $\bar{\mathcal{D}}$. The idea is that, under the assumption that points with the same label are sufficiently similar for the subgroup transformation to account for the important difference\footnote{Keeping MNIST hand-written digits as our paradigmatic example, digits with the same label differ by small transformations that account for handwriting style differences.}, we can use labels to group data points, and compare those data points using methods such as the one presented in this paper. We leave the fully unsupervised symmetry detection setting for future work although we will emphasize that the proposed method can, in principle, be used in such setting without substantial changes to the architecture. 

\medskip
\section{Experiments}\label{sec:experiments}
\subsection{Experiment setting}
In practice, we experiment with a dataset of MNIST digits transformed with either 2D rotations or translations in one direction. To test the method's ability to learn distributions of these transformations, for each one-parameter subgroup $\phi\in\{SO(2),T(2)\}$ we construct a dataset $\left\{x_i,T(x_i,t_i)\right\}_{i=1}^{N}$ by sampling the parameters $t_i\in\mathbb{R}$ from various \emph{multimodal} distributions.

As in \cite{rao1998learning}, the dataset is composed of signals $I: M \longrightarrow \mathbb{R}$ regularly-sampled from a discrete grid of $n^2$ points $(x,y)\in\mathbb{R}^2$ for $n=28$. The signals $I$ are vectorised into points in $\mathbb{R}^{784}$ as described in Section \ref{subsec:discretegen}. The implementation of the naive model is available \href{https://github.com/victoria-klein/learning-lie-group-symmetries.git}{here}.

\subsection{Main model experiments}
The naive model architecture outlined in \ref{subsubsec:mainmodel} consists of a fully-connected, 3-layer MLP for $f_\theta$ that was trained jointly with the coefficients $\alpha_{ij}$ using Adam \cite{kingma2014adam} with a learning rate of 0.001. Given the disproportionate number of trainable parameters in $f_\theta$ and the 6 coefficients in $\alpha$, updating $\alpha_{ij}$ roughly 10 times for every update of $\theta$ in $f_\theta$ was found to be beneficial during training.

\begin{figure}[h]
  \centering
  \subfigure[Rotation]{\includegraphics[width=\linewidth]{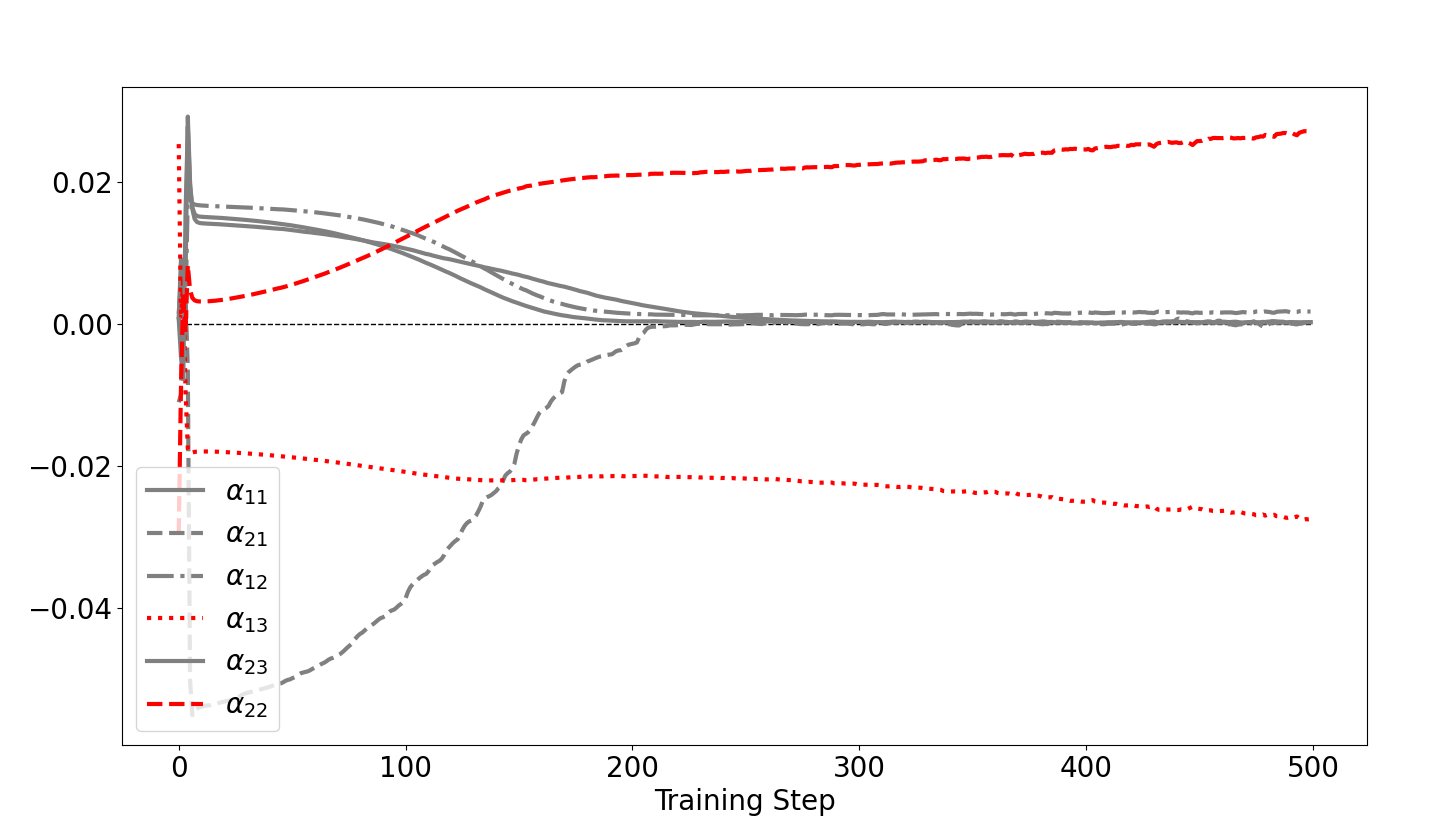}}\label{fig:subplot1}
  \hfill
  \subfigure[Translation in $x$]{\includegraphics[width=\linewidth]{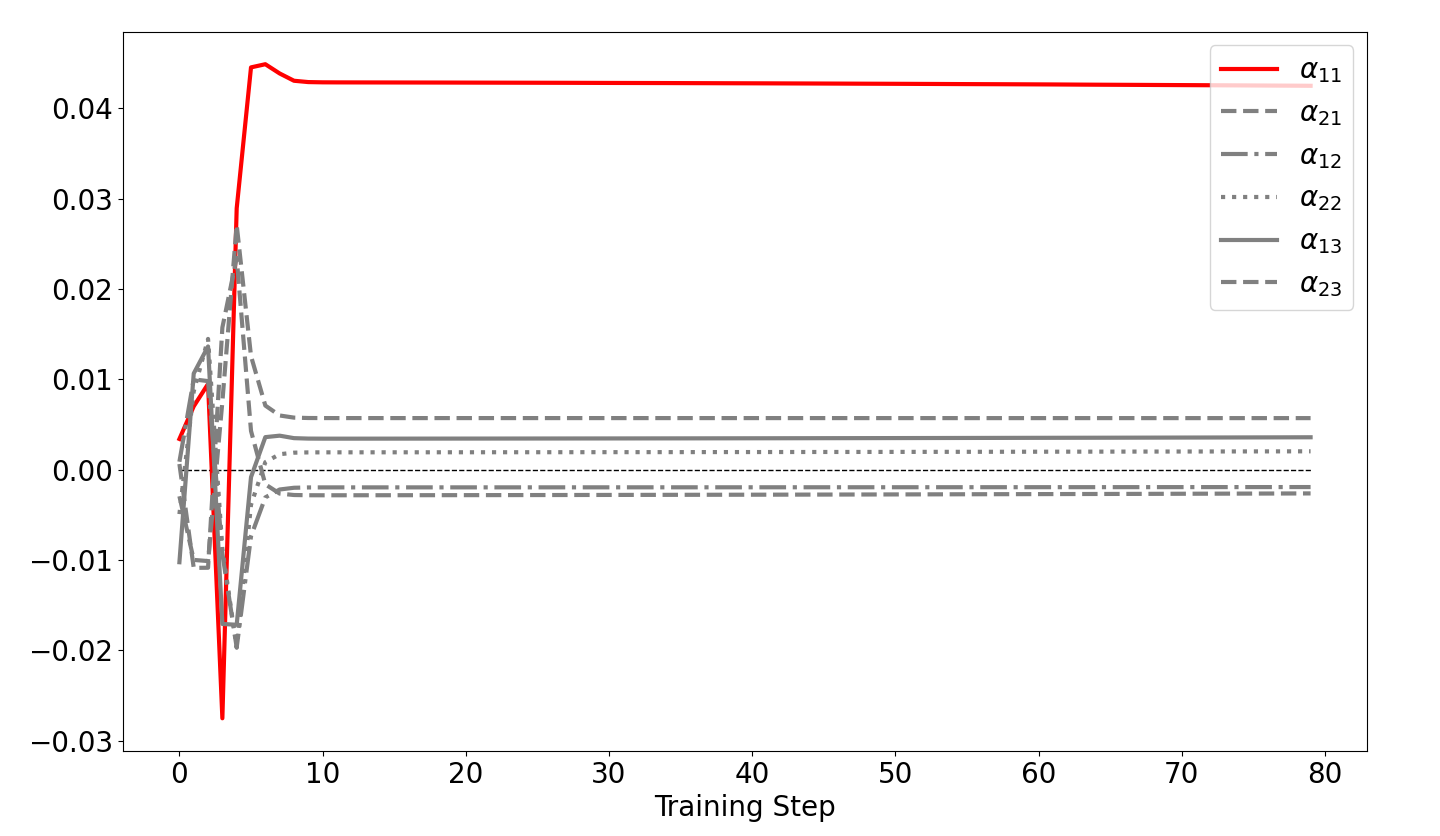}\label{fig:subplot2}}
  \caption{Training evolution of the coefficients $\alpha$ defining the generator $L^{\alpha}$ of the one-parameter subgroup, that are shown to converge to the ground-truth non-zero coefficients $\alpha$ for rotated ($-\alpha_{22}=\alpha_{13}=1$ and $0$ otherwise) and translated ($\alpha_{11}=1$ and $0$ otherwise) MNIST.}
  \label{fig:main}
\end{figure}

\paragraph{Coefficients} Figure \ref{fig:main} shows the evolution of $\alpha_{ij}$ during training. It can be seen that after a few hundred steps, the coefficients $\alpha_{ij}$ that do not correspond to the infinitesimal generator of the symmetry expressed by the dataset drop to zero, while those that do, settle to values compatible with those of the ground truth generator $L$.

\subsection{Latent model experiments}
The latent model outlined in \ref{subsubsec:latentmodel} consists of a fully-connected, 3-layer MLP $f_\theta$, as in \eqref{eq:modelapprox}, to approximate $\hat{t}$, and two fully-connected, 3-layer MLPs with decreasing/increasing hidden dimensions for the encoder $h_\psi$ and $d_\psi$. We set the latent space to $n_Z=25$. Similar to the naive model experiment above, $f_\theta$ was trained jointly with the coefficients $\alpha_{ij}$ using Adam \cite{kingma2014adam} with learning rate 0.001.

\begin{figure}
  \centering
  \subfigure[Unimodal distribution]{\includegraphics[width=0.8\linewidth]{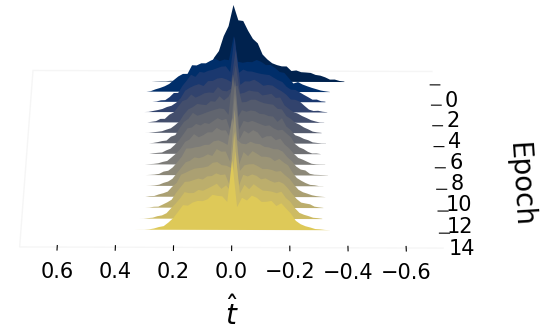}}\label{fig:subplot1}
  \hfill
  \subfigure[Bimodal distribution]{\includegraphics[width=0.8\linewidth]{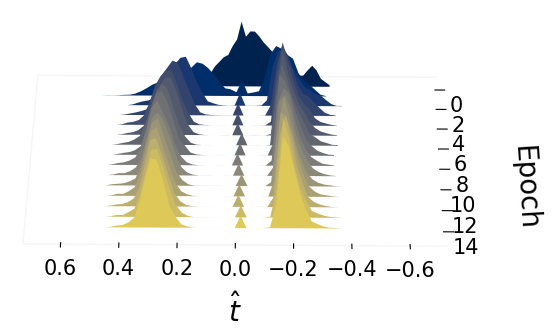}}\label{fig:subplot2}
  \hfill
  \subfigure[3-mode distribution]{\includegraphics[width=0.8\linewidth]{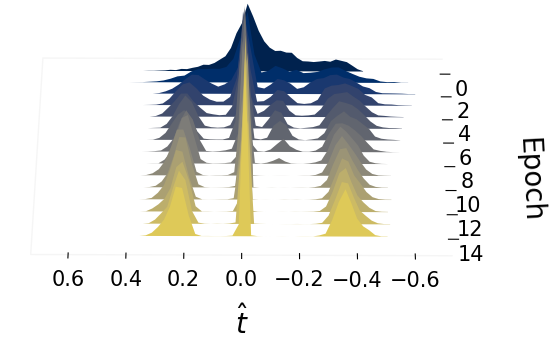}}\label{fig:subplot3}
  \hfill
  \subfigure[5-mode distribution]{\includegraphics[width=0.8\linewidth]{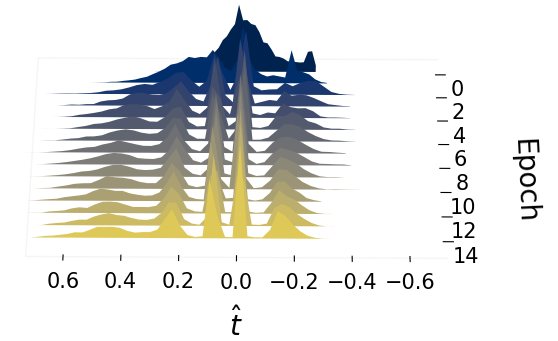}}\label{fig:subplot3}
  \caption{Training evolution of the distributions $p(\hat{t})$ of the learned parameters $\hat{t}$ computed by $f_\theta$ for the validation set. The figure shows that $p(\hat{t})$ resembles the original multi-modal distributions $p(t)$ of the transformations expressed by the dataset.}
  \label{fig:latent}
\end{figure}

\paragraph{Parameters}
After every epoch (roughly 500 steps), the outputs of $\hat{t}=f_\theta$ were collected in a histogram to show $p(\hat{t})$. Figure \ref{fig:latent} shows how the distribution of $\hat{t}$ changes during training and how multimodal distributions are clearly recovered, showing the same number of modes as the ground truth distribution from which the transformations were sampled.

\section{Related Work}\label{sec:related}
\textbf{Symmetries in Neural Networks}
Numerous studies have tackled the challenges associated with designing neural network layers and/or models that are equivariant with respect to specific transformations \citep{finzi2021practical}. These transformations include continuous symmetries such as scaling \citep{worrall2019deep}, rotation on spheres \citep{cohen2018spherical}, local gauge transformations \citep{cohen2019gauge} and general E(2) transformations on the Euclidean plane \citep{weiler2019general}, as well as discrete transformations like permutations of sets \citep{zaheer2017deep} and reversing symmetries \citep{valperga2022learning}. Another line of research focuses on establishing theoretical principles and practical techniques for constructing general group-equivariant neural networks. Research in such areas show improved performances on tasks related to symmetries, but nonetheless require prior knowledge about the symmetries themselves.
\\
\\
\textbf{Symmetry Detection}
Symmetry detection aims to discover symmetries from observations, a learning task that is of great importance in of itself. Detecting symmetries in data not only lends itself to more efficient and effective machine learning models but also in discovering fundamental laws that govern data, a long-standing area of interest in the physical sciences. Learned symmetries can then be incorporated after training in equivariant models or used for data augmentation for downstream tasks. In physics and dynamical systems, the task of understanding and discovering symmetries is a crucial one; in classical mechanics and more generally Hamiltonian dynamics, continuous symmetries of the Hamiltonian are of great significance since they are associated, through Noether's theorem \citep{Noether1918}, to conservation laws such as conservation of angular momentum or conservation of charge.

 The first work on learning symmetries of one-parameter subgroups from observations were \citet{rao1998learning} and \citet{miao2007learning}, which outline MAP-inference methods for learning infinitesimally small transformations. \citet{sohl2010unsupervised} propose a transformation-specific smoothing operation of the transformation space to overcome the issue of a highly non-convex reconstruction objective that includes an exponential map. These methods are close to ours in that we also make use of the exponential map to obtain group elements from their Lie algebra. Despite this, \citet{sohl2010unsupervised} do not consider the task of characterizing the distribution of the parameter of the subgroup nor do they consider the whole of pixel-space, using small patches instead. \citet{cohen2014learning} focus on disentangling and learning the distributions of multiple compact ``toroidal" one-parameter subgroups in the data.

 \textbf{Neural Symmetry Detection} A completely different approach to symmetry discovery is that of \citet{sanborn2022bispectral}, who's model uses a group invariant function known as the bispectrum to learn group-equivariant and group-invariant maps from observations. \citet{benton2020learning} consider a task similar to ours, attempting to learn groups with respect-to-which the data is invariant, however, the objective places constraints directly on the network parameters as well as the distribution of transformation parameters with which the data is augmented. Alternatively, \citet{dehmamy2021automatic} require knowledge of the specific transformation parameter for each input pair (differing by that transformation), unlike our model, where no knowledge of the one-parameter group is used in order to find the distribution of the transformation parameter.

 \textbf{Latent Transformations} Learning transformations of a one-parameter subgroup in latent space (whether that subgroup be identical to the one in pixel space or not) has been accomplished by \citet{keurti2023homomorphism} and \citet{zhu2021commutative}. Nevertheless, other works either presuppose local structure in the data by using CNNs instead of fullly-connected networks or focus on disentangling interpretable features instead of directly learning generators that can be used as an inductive bias for a new model.

In contrary to the other works mentioned above, we propose a promising framework in which we can simultaneously
\begin{itemize}
    \item perform symmetry detection in pixel-space, without assuming any inductive biases are present in the data \textit{a priori},
    \item parametrize the generator such that non-compact groups (e.g. translation) can be naturally incorporated,
    \item and learn both the generator and the parameter distributions.
\end{itemize}

\section{Discussion}
In this work we proposed a framework for learning one-parameter subgroups of Lie group symmetries from observations. Our method uses a neural network to predict the one-parameter of every transformation that has been applied to datapoints, and the coefficients of a linear combination of pre-specified generators. We show that our method can learn the correct generators for a variety of transformations as well as characterize the distribution of the parameter that has been used for transforming the dataset.

% A few limitations need to be discussed: for \emph{truly} unsupervised learning of symmetries based only on class labels two points that belong to the same class must be, up to the Lie group transformation, to be sufficiently similar for the Lie group transformation to account for the important difference. For MNIST, two digits with the same label differ because of handwriting. As explained in Section \ref{sec:method} in this work, in line with other works on symmetry discovery, we consider only the case of input pairs that only differ by exact Lie group transformations. We stress that our method can in principle be applied to the more general setting and leave this for future work.

While the goal of learning both the coefficients of the generator and the distribution of the transformation parameter has not been accomplished by only one model in this work, modifying our existing framework to do so is a priority for future work. In addition, the proposed method lends itself well to being composed to form multiple layers, which can then be applied to datasets that express multiple symmetries. By doing so, ideally, each layer would learn one individual symmetry. We leave this study, and the more general, fully unsupervised setting described in \ref{sec:generalsetting}, for future work.

\section*{Acknowledgements}
This publication is based on work partially supported by the EPSRC Centre for Doctoral Training in Mathematics of Random Systems: Analysis, Modelling and Simulation (EP/S023925/1) and the Dorris Chen Award granted by the Department of Mathematics, Imperial College London.
\nocite{langley00}

\bibliography{example_paper}
\bibliographystyle{icml2023}

%%%%%%%%%%%%%%%%%%%%%%%%%%%%%%%%%%%%%%%%%%%%%%%%%%%%%%%%%%%%%%%%%%%%%%%%%%%%%%%
%%%%%%%%%%%%%%%%%%%%%%%%%%%%%%%%%%%%%%%%%%%%%%%%%%%%%%%%%%%%%%%%%%%%%%%%%%%%%%%
% APPENDIX
%%%%%%%%%%%%%%%%%%%%%%%%%%%%%%%%%%%%%%%%%%%%%%%%%%%%%%%%%%%%%%%%%%%%%%%%%%%%%%%
%%%%%%%%%%%%%%%%%%%%%%%%%%%%%%%%%%%%%%%%%%%%%%%%%%%%%%%%%%%%%%%%%%%%%%%%%%%%%%%
% \newpage
% \appendix
% \onecolumn

% You can have as much text here as you want. The main body must be at most $8$ pages long.
% For the final version, one more page can be added.
% If you want, you can use an appendix like this one, even using the one-column format.
%%%%%%%%%%%%%%%%%%%%%%%%%%%%%%%%%%%%%%%%%%%%%%%%%%%%%%%%%%%%%%%%%%%%%%%%%%%%%%%
%%%%%%%%%%%%%%%%%%%%%%%%%%%%%%%%%%%%%%%%%%%%%%%%%%%%%%%%%%%%%%%%%%%%%%%%%%%%%%%

\end{document}